\documentclass[runningheads]{llncs}
\usepackage{graphicx}
\usepackage{amsmath}
\usepackage{algorithm}
\usepackage{algorithmic}
\usepackage{graphicx}
\usepackage{booktabs}
\usepackage{multirow}
\usepackage{subcaption}
\usepackage{xspace}
\usepackage{siunitx}
\usepackage{amsmath}    
\usepackage[dvipsnames]{xcolor}
\usepackage{colortbl}
\usepackage{amssymb}   
\usepackage{bm}       
\usepackage{caption}
\usepackage{cleveref}
\usepackage{wrapfig}
\newtheorem{defn}{Definition}
\newtheorem{remk}{Remark}

\begin{document}
\title{Divide, Weight, and Route: \\
Difficulty-Aware Optimization with Dynamic Expert Fusion 
for Long-tailed Recognition}

\titlerunning{Divide, Weight, and Route}
\author{Xiaolei Wei\inst{1} \and
        Yi Ouyang\inst{2} \and
        Haibo Ye\inst{1}\thanks{Corresponding author: Haibo Ye (email: yhb@nuaa.edu.cn)}}
\authorrunning{X. Wei et al.}
\institute{ the MIIT Key Laboratory of Pattern Analysis and Machine Intelligence, Collaborative Innovation Center of Novel Software Technology and Industrialization, College of Computer Science and Technology, Nanjing University of Aeronautics and Astronautics, Nanjing, China\\
\email{\{xiaolei\_wei,yhb\}@nuaa.edu.cn} \and
Shanghai Electro-Mechanical Engineering Institute, Shanghai 200240,China National Key Laboratory on Automatic Target Recognition, Shanghai 200240, China\\
\email{Ricardo\_ouyang@163.com}}

\maketitle              
\begin{abstract}
Long-tailed visual recognition is challenging not only due to class imbalance but also because of varying classification difficulty across categories. Simply reweighting classes by frequency often overlooks those that are intrinsically hard to learn. To address this, we propose \textbf{DQRoute}, a modular framework that combines difficulty-aware optimization with dynamic expert collaboration. DQRoute first estimates class-wise difficulty based on prediction uncertainty and historical performance, and uses this signal to guide training with adaptive loss weighting. On the architectural side, DQRoute employs a mixture-of-experts design, where each expert specializes in a different region of the class distribution. At inference time, expert predictions are weighted by confidence scores derived from expert-specific OOD detectors, enabling input-adaptive routing without the need for a centralized router. All components are trained jointly in an end-to-end manner. Experiments on standard long-tailed benchmarks demonstrate that DQRoute significantly improves performance, particularly on rare and difficult classes, highlighting the benefit of integrating difficulty modeling with decentralized expert routing.
\end{abstract}

\section{Introduction}
\label{sec:introduction}

Visual recognition systems deployed in real-world environments frequently encounter long-tailed data distributions~\cite{shao2023identity,shao2024joint,hsr,shrec}, where a small number of head classes dominate the training set, while the majority of tail classes remain severely underrepresented. This imbalance biases deep neural networks toward head classes, leading to poor generalization on rare but often critical categories~\cite{pmlr-v235-li24bx}. The consequences of this issue are particularly severe in safety-critical domains such as autonomous driving, where rare hazards must be reliably detected, and medical imaging, where uncommon pathologies require accurate identification.

\begin{figure}
    \centering
    \includegraphics[width=\linewidth]{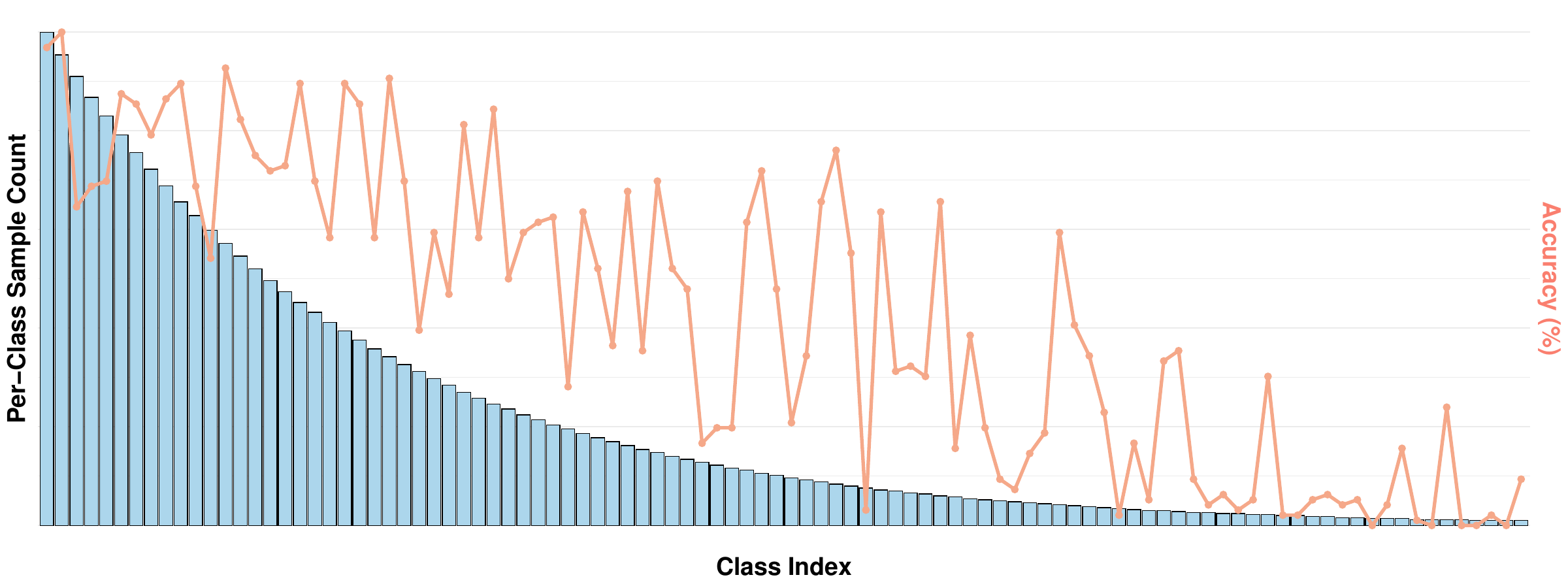}
    \caption{\textbf{Class-wise accuracy vs. training sample count on CIFAR-100-LT (Imbalance Ratio=100).} 
        Accuracy does not monotonically correlate with class size, indicating that quantity-based methods alone are insufficient to address long-tailed recognition challenges.}
    \label{fig:quantity_vs_performance}
\end{figure}

To mitigate the effects of class imbalance, most existing approaches focus on modeling class \textit{quantity}. Techniques such as resampling, loss reweighting, and decoupled training modify the learning process based on class frequency. However, these methods implicitly assume that class frequency is a reliable proxy for learning difficulty. As shown in Fig.\ref{fig:quantity_vs_performance}, this assumption often fails. On CIFAR-100-LT, class-wise accuracy exhibits only a weak correlation with sample count: some rare classes are learned well, while certain moderately frequent classes perform poorly. These observations suggest that \textit{learning difficulty} is influenced not only by data scarcity but also by class-level complexity, which quantity-based methods overlook.

To address this limitation, we introduce a difficulty-aware learning strategy that complements class frequency with a direct estimate of class-level difficulty. We define difficulty using two test-time signals: class-wise accuracy and logit entropy. These metrics jointly capture how confidently and correctly the model classifies each class under realistic inference conditions. Classes with low accuracy and high entropy are considered more difficult and are assigned greater importance through dynamic loss reweighting. This allows the model to allocate more learning capacity to classes that are inherently ambiguous or underperforming, regardless of their frequency in the training set, resulting in more balanced representation learning.

In parallel, we propose a dynamic expert fusion mechanism designed to handle the distributional heterogeneity across the class spectrum. Building on the mixture-of-experts paradigm, we train three experts, each specialized in a different subset of classes: one on all classes, one on medium and tail classes, and one on tail classes only. Rather than aggregating expert outputs uniformly, we introduce a dynamic routing strategy that adaptively fuses expert predictions based on the input's distributional alignment. This routing is guided by an out-of-distribution (OOD) scoring mechanism, which estimates how well an input aligns with the training distribution of each expert. The resulting soft routing enables the model to emphasize more appropriate experts for each input, enhancing robustness and specialization without requiring hard expert selection.

We propose \textbf{DQRoute}, a unified framework that integrates \textbf{D}ifficulty and \textbf{Q}uantity-aware optimization with dynamic expert \textbf{routing} for long-tailed visual recognition. During training, \textbf{DQRoute} reweights class contributions based on both estimated difficulty and frequency. At inference, it dynamically routes inputs through distribution-specialized experts using learned routing weights derived from their distributional fit. This dual strategy, which optimizes for hard classes and routes inputs to specialized experts, enables \textbf{DQRoute} to consistently outperform strong baselines and recent state-of-the-art methods across multiple long-tailed benchmarks.

Our main contributions are summarized as follows:
\begin{itemize}
\item We propose \textbf{DQRoute}, a unified framework for long-tailed recognition that jointly addresses both class quantity imbalance and class difficulty, supported by a theoretical analysis of difficulty-gradient interactions.

\item We introduce a difficulty-aware loss reweighting strategy that dynamically adjusts class importance based on class-wise accuracy and entropy, allowing the model to focus on underperforming and ambiguous classes during training.

\item We develop a dynamic expert fusion mechanism that adaptively combines the outputs of distribution-specialized experts using input-specific distributional alignment, significantly outperforming static or uniform fusion schemes.
\end{itemize}


\section{Related Works}
\label{sec:works}

Long-tailed visual recognition remains a fundamental challenge in real-world applications due to the heavy imbalance in category frequencies. Existing solutions can be broadly categorized into two directions: (1) loss-level or sample-level strategies that modify the training objective or data distribution, and (2) architecture-level approaches that introduce modular structures to handle data heterogeneity. We briefly review the most relevant works in each direction.

\subsection{Loss and Data-Level Approaches}

A large body of work focuses on mitigating long-tailed bias through loss function design, data augmentation, re-sampling, and contrastive learning.

\noindent \textbf{Re-sampling and Loss design.} Re-sampling methods~\cite{kang2019decoupling,wang2020devil,zang2021fasa,ren2020balanced,feng2021exploring} dynamically adjust sampling rates to reduce bias. Balanced loss functions aim to reduce the dominance of head classes during optimization. Focal Loss~\cite{lin2017focal}, CB Loss~\cite{cui2019class}, LDAM~\cite{cao2019learning}, and Seesaw Loss~\cite{wang2021seesaw} modify the decision boundaries or gradient contributions to favor tail classes. Equalization Loss~\cite{tan2020equalization,tan2021equalization} and AREA~\cite{chen2023area} further enhance balance through adaptive gradient reweighting and region-aware loss terms. Logits adjustment methods~\cite{wu2021adversarial,hong2021disentangling,zhang2021distribution,lione2025} aim to adjust models’ output logits to enhance the differentiation between head and tail classes. Difficulty-aware losses~\cite{sinha2020class,yu2022re,sinha2023difficulty,son2025difficulty} re-weight samples based on prediction uncertainty or estimated learning difficulty to emphasize ambiguous or under-performing classes.

\noindent \textbf{Data augmentation.} Mixup-based methods~\cite{zhang2017mixup,chou2020remix,gao2022dynamic,shao2025mol} and MetaSAug~\cite{li2021metasaug,li2025hybrid} synthesize new samples for minority classes to improve generalization. Generative approaches like LTGC~\cite{zhao2024ltgc} use large language models to augment tail data.

\noindent \textbf{Contrastive learning.} Prototype-based contrast~\cite{zhu2022balanced}, parametric contrast~\cite{cui2021parametric}, consistency-aware contrast~\cite{du2023global}, and hybrid contrastive methods~\cite{wang2021contrastive} have been proposed to improve feature discrimination in the presence of class imbalance.

\subsection{Architecture-Level Approaches}

Another line of work explores modular architectures to decompose the long-tailed learning problem across multiple specialized components.

\noindent \textbf{Mixture-of-Experts (MoE).} BBN~\cite{zhou2020bbn} separate representation learning from classifier optimization. LFME~\cite{xiang2020learning} trains experts on distinct data segments and consolidates their knowledge into a unified student model. Modular expert frameworks such as RIDE~\cite{wang2020long} and ACE~\cite{cai2021ace} assign different experts to different regions of the label space, dynamically combining their output during inference. SADE~\cite{zhang2022self} and EME~\cite{bai2024eme} further improve expert diversity and fusion through self-supervised and energy-based techniques. MDCS~\cite{zhao2023mdcs} enhances robustness by enforcing consistency or stepwise knowledge transfer among experts. Recent works like LTRL~\cite{zhao2024ltrl} and ResLT~\cite{cui2022reslt} focus on adapting expert confidence or residual pathways to better model head-tail differences. OpenworldAUC~\cite{openworldauc2025} introduces a novel metric that remains insensitive to class imbalance, ensuring consistent evaluation across both known and unknown categories. Building on this metric, it further proposes a prompt-based MoE framework to improve the overall imbalance open-world recognition.

\noindent \textbf{OOD-aware routing.} Several methods~\cite{wang2022partial,wei2024eat,bai2023effectiveness,miao2024out,he2025long} integrate OOD (out-of-distribution) detection with long-tailed visual recognition. DQRoute (ours) uses expert-specific OOD confidence to drive decentralized routing without a centralized router.


\section{Methodology}
\label{sec:method}
\begin{figure}[t]
    \centering
    \includegraphics[width=\linewidth]{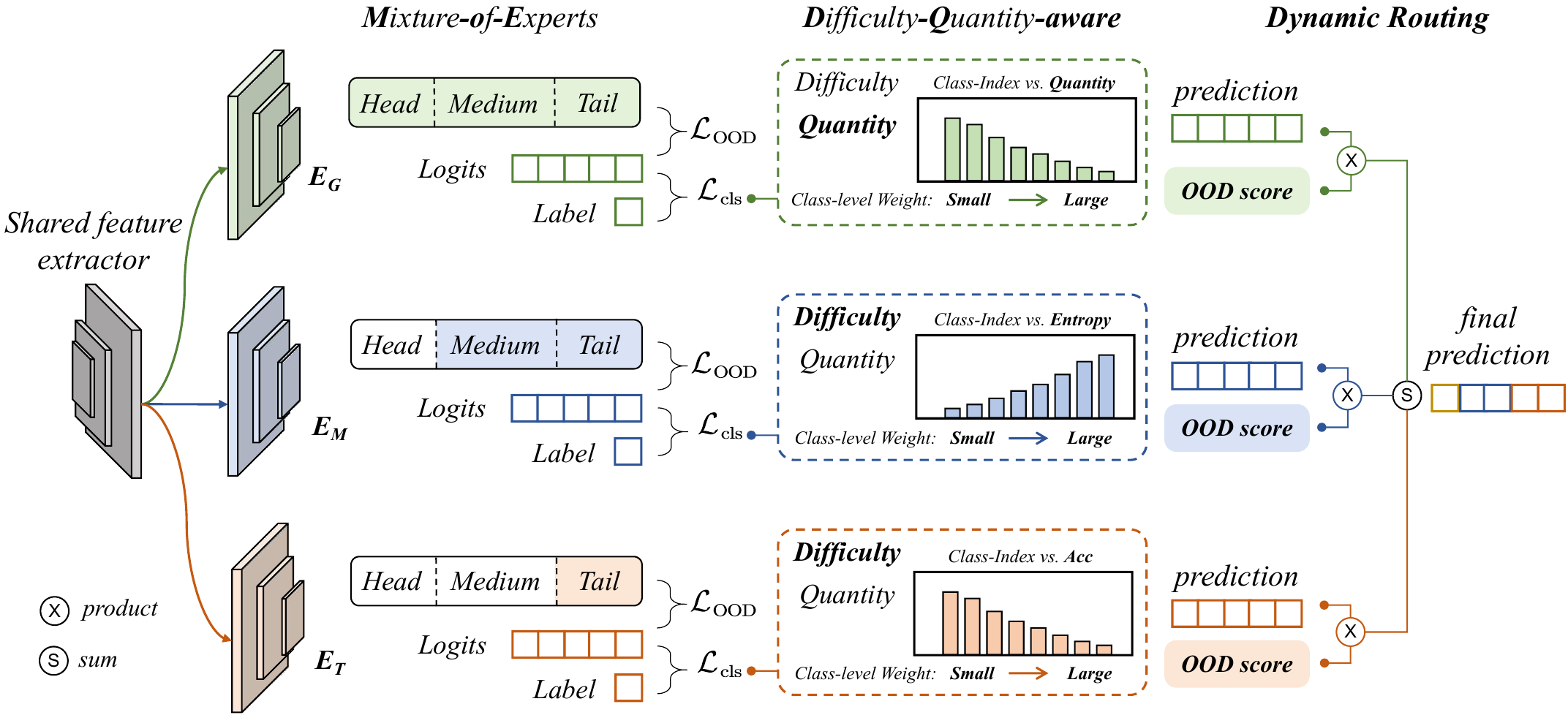}
    \caption{
        \textbf{DQRoute Architecture.}
        (1) Input image is processed by a shared feature extractor. 
        (2) Output featuress are passed to three distribution-specialized experts. 
        (3) Each expert produces both a class prediction, optimized by a classification loss reweighted by class difficulty and quantity.
        (4) Each expert produces an OOD score, which is used to fuse expert predictions into the final prediction.
    }
    \label{fig:architecture}
\end{figure}

We present \textbf{DQRoute}, a unified framework designed to address the challenges of long-tailed visual recognition by jointly incorporating \textit{difficulty-aware optimization} and \textit{dynamic expert routing}. The core idea is to not only acknowledge the long-tail distribution in terms of class frequency, but also explicitly model the inherent difficulty of each class and dynamically route samples to specialized experts based on their distributional characteristics.

As illustrated in Fig.\ref{fig:architecture}, DQRoute consists of four main components: a shared feature extractor, a difficulty-aware reweighting mechanism, a set of distribution-specialized experts, and an expert-driven routing module. An input image is first processed by a shared backbone (e.g., ResNet-32), which extracts a feature representation. This feature is then passed to three parallel experts, each trained on a specific subset of the class distribution (head, medium, or tail). Each expert produces both a class prediction and a confidence score via its internal OOD detection head. These confidence scores are normalized to produce routing weights that determine how much each expert contributes to the final prediction. This decentralized, confidence-driven routing strategy allows the model to adaptively combine the strengths of specialized experts based on the input characteristics.

\subsection{Difficulty-Aware Optimization}
\label{subsec:difficulty}
Reweighting or resampling based on class frequency is a common approach in long-tailed recognition. However, this conflates \textit{rarity} with \textit{difficulty}, assuming underrepresented classes are inherently harder to learn. As shown in Fig.\ref{fig:quantity_vs_performance}, classes with similar frequencies can exhibit diverse uncertainties, indicating that difficulty is not solely frequency-driven.

To address this, we propose a data-driven difficulty measure that combines \textit{prediction ambiguity} and \textit{classification performance}:
\begin{defn}[Class Difficulty]
Let $\mathcal{H}_c$ be the average entropy and $\mathcal{A}_c$ the exponentially smoothed accuracy for class $c$. The difficulty score $d_c \in [0, 1]$ is defined as:
\begin{equation}
d_c = \frac{\mathcal{H}_c}{\mathcal{H}_{\max}} + \lambda \left(1 - \frac{\mathcal{A}_c}{\mathcal{A}_{\max}}\right),
\end{equation}
with $\lambda$ balancing the two components.
\end{defn}

\begin{remk}
$d_c$ captures both prediction uncertainty and performance. High $d_c$ indicates either high entropy or low accuracy, making it more adaptive than frequency-based heuristics.
\end{remk}
We integrate $d_c$ into training via class-wise dynamic reweighting. The difficulty-based weight at epoch $t$ is:
\begin{equation}
w_c^{(t)} = \frac{w_c^{(t-1)} \cdot \exp(\gamma d_c)}{\sum_{j=1}^{C} w_j^{(t-1)} \cdot \exp(\gamma d_j)},
\end{equation}
where $\gamma$ controls adaptation sharpness. This emphasizes difficult classes without destabilizing training.

To balance empirical difficulty with prior knowledge from class frequency, we define the final weight as:
\begin{equation}
\tilde{w}_c^{(t)} = \alpha \cdot w_c^{(t)} + (1 - \alpha) \cdot q_c,
\end{equation}
where $q_c$ is the normalized quantity-based weight and $\alpha \in [0,1]$ balances the two. This unified scheme leverages both model uncertainty and data imbalance, offering a flexible and effective strategy for long-tailed optimization.

\subsection{Mixture-of-Experts with Dynamic Routing}
\label{subsec:experts}

\noindent \textbf{Mixture-of-Experts.}
To handle diverse class distributions and complexities, we adopt a modular architecture with three specialized experts: a general expert $E_G$, a medium-shot expert $E_M$, and a tail expert $E_T$. All share a feature extractor $\phi(\cdot)$ but have independent classification heads. $E_G$ is trained on the full dataset $\mathcal{D}$, $E_M$ on classes with fewer than $\tau_m$ samples, and $E_T$ on those below $\tau_t < \tau_m$.

Specifically, each expert $E_k$ is trained with a standard class-conditional cross-entropy loss:
\begin{equation}
\mathcal{L}_{\text{cls}}^{(k)} = \mathbb{E}_{(x, y) \sim \mathcal{D}_k} \left[ \mathcal{L}_{\text{CE}}(f_{E_k}(x), y) \right],
\end{equation}
where $\mathcal{D}_k$ contains samples from class group $\mathcal{C}_k$.

\noindent \textbf{Dynamic Routing.}
To combine expert outputs, we use a decentralized routing mechanism based on expert-specific confidence. Each expert $E_k$ includes an OOD head that outputs a confidence score $s_k(x) \in [0,1]$, which is normalized into routing weights:
\begin{equation}
\alpha_k(x) = \frac{s_k(x)}{\sum_{j=1}^{3} s_j(x)},
\end{equation}
\begin{equation}
P(y \mid x) = \sum_{k=1}^{3} \alpha_k(x) \cdot f_{E_k}(x).
\end{equation}
To supervise the OOD heads, we define a binary label $b_k(x) = \mathbb{I}[y \in \mathcal{C}_k]$ and apply binary cross-entropy loss:
\begin{equation}
\mathcal{L}_{\text{OOD}}^{(k)} = - \left[ b_k(x) \log s_k(x) + (1 - b_k(x)) \log (1 - s_k(x)) \right],
\end{equation}
\begin{equation}
\mathcal{L}_{\text{OOD}} = \sum_{k=1}^{3} \mathcal{L}_{\text{OOD}}^{(k)}.
\end{equation}
This routing strategy enables each expert to assess its own relevance and dynamically contribute to final predictions. Further analysis of $\mathcal{L}_{\text{OOD}}$ design is provided in Section~\ref{subsec:ablation}.

\subsection{Joint Learning Objective}
\label{subsec:training}

\noindent \textbf{DQRoute} integrates difficulty-aware optimization, distribution-specialized experts, and dynamic routing into a unified training framework. All components are trained jointly to promote specialization, adaptability, and robust performance across the long-tailed label space.

The total loss consists of two parts: (1) a classification loss $\mathcal{L}_{\text{cls}}$ that incorporates both expert relevance and class difficulty, and (2) an OOD detection loss $\mathcal{L}_{\text{OOD}}$ that supervises expert confidence estimation (Section~\ref{subsec:experts}).
\begin{equation}
\mathcal{L}_{\text{total}} = \mathcal{L}_{\text{cls}} + \lambda_{\text{OOD}} \cdot \mathcal{L}_{\text{OOD}},
\end{equation}
where $\lambda_{\text{OOD}}$ is a balancing hyperparameter. The classification loss is computed as a weighted ensemble over expert predictions:
\begin{equation}
\mathcal{L}_{\text{cls}} = \mathbb{E}_{(x, y) \sim \mathcal{D}} \left[ \sum_{k=1}^{3} \alpha_k(x) \cdot \tilde{w}_y \cdot \mathcal{L}_{\text{CE}}(f_{E_k}(x), y) \right],
\end{equation}
where $\alpha_k(x)$ is the routing weight from expert $E_k$, and $\tilde{w}_y$ is the final class weight that combines both difficulty and quantity (as defined in Section~\ref{subsec:difficulty}).

This joint objective enables the model to dynamically allocate learning effort based on both expert confidence and class-level difficulty, resulting in improved performance across all regions of the long-tailed distribution.

\section{Experiments}
\label{sec:experiments}

\subsection{Datasets and Implementation Details}
\label{subsec:setup}

\textbf{Datasets.} We evaluate our method on four widely used long-tailed benchmarks. \textbf{CIFAR-10-LT} and \textbf{CIFAR-100-LT}~\cite{krizhevsky2009learning} are derived from the original CIFAR datasets using an exponential sampling strategy with imbalance ratios (IR) of 10, 50, and 100. \textbf{ImageNet-LT}~\cite{Liu2019Large} is a naturally imbalanced subset of ImageNet with 115,846 images across 1,000 classes. \textbf{Places-LT}~\cite{Liu2019Large} is a long-tailed version of the Places-365 scene dataset, with 62,500 images and a severe class imbalance (from 4,980 to 5 samples per class). These benchmarks span both object-level and scene-level recognition under varying levels of imbalance.

\noindent \textbf{Implementation Details.} We follow the experimental setup of SADE~\cite{zhang2022self}. ResNet-32 is used for CIFAR, ResNeXt-50 for ImageNet-LT, and ResNet-152 for Places-LT. Unless otherwise noted, all models are trained for 200 epochs using SGD with a momentum of 0.9, weight decay of $5 \times 10^{-4}$, and a learning rate linearly decaying from 0.1. Standard data augmentations (random cropping and flipping) are applied. Following prior works~\cite{cai2021ace,zhang2022self,zhao2024ltrl}, we report results on a balanced test set and include accuracy across \textbf{many}-shot ($>100$ images), \textbf{medium (mid.)}-shot (20–100), and \textbf{few}-shot ($<20$) classes.

\subsection{Comparison with State-of-the-Art}
\label{subsec:main_results}

DQRoute achieves state-of-the-art or highly competitive performance on both tail and head classes, across various datasets and imbalance settings. Its effectiveness among difficulty-based and multi-expert models demonstrates its robustness and versatility.

\begin{table}[t]
  \centering
  \caption{Quantitative comparison on CIFAR-100-LT-IR100. Best results are shown in \textbf{bold}. A \textcolor{ForestGreen}{($\cdot$)} indicates the increase from the CE baseline, and a \textcolor{red}{($\cdot$)} indicates the decrease from the CE baseline.}
    \begin{tabular}{c|c|c|c|c|c}
    \toprule
    \multicolumn{1}{c|}{\multirow{2}[4]{*}{Method}} & \multirow{2}[4]{*}{Multi-Experts} & \multicolumn{4}{c}{Accuracy} \\
\cmidrule{3-6}    \multicolumn{1}{c|}{} &       & All   & Many  & Medium  & Few  \\
    \midrule
    CE    &       & 41.4  & 66.1  & 37.3  & 10.6 \\
    OLTR~\cite{Liu2019Large}  &       & 41.2 \textcolor{red}{(-0.2)} & 61.8 \textcolor{red}{(-4.3)} & 41.4 \textcolor{ForestGreen}{(+4.1)} & 17.6 \textcolor{ForestGreen}{(+7.0)} \\  
    $\tau$-norm~\cite{kang2019decoupling} &       & 43.2 \textcolor{ForestGreen}{(+1.8)} & 65.7 \textcolor{red}{(-0.4)} & 43.6 \textcolor{ForestGreen}{(+6.3)} & 17.3 \textcolor{ForestGreen}{(+6.7)} \\
    cRT~\cite{kang2019decoupling}   &       & 43.3 \textcolor{ForestGreen}{(+1.9)} & 64.4 \textcolor{red}{(-1.7)} & 44.8 \textcolor{ForestGreen}{(+7.5)} & 18.1 \textcolor{ForestGreen}{(+7.5)} \\      
    BBN~\cite{zhou2020bbn}   & \checkmark     & 39.4 \textcolor{red}{(-2.0)} & 47.2 \textcolor{red}{(-18.9)} & 49.4 \textcolor{ForestGreen}{(+12.1)} & 19.8 \textcolor{ForestGreen}{(+9.2)} \\
    LFME~\cite{xiang2020learning}  & \checkmark     & 43.6 \textcolor{ForestGreen}{(+2.2)} & -     & -     & - \\
    ACE~\cite{cai2021ace}   & \checkmark     & 49.4 \textcolor{ForestGreen}{(+8.0)} & 66.1 \textcolor{ForestGreen}{(+0.0)} & \textbf{55.7 \textcolor{ForestGreen}{(+18.4)}} & 23.5 \textcolor{ForestGreen}{(+12.9)} \\
    RIDE~\cite{wang2020long}  & \checkmark     & 48.0 \textcolor{ForestGreen}{(+6.6)} & \textbf{67.4 \textcolor{ForestGreen}{(+1.3)}} & 49.5 \textcolor{ForestGreen}{(+12.2)} & 23.7 \textcolor{ForestGreen}{(+13.1)} \\
    SADE~\cite{zhang2022self}  & \checkmark     & 49.4 \textcolor{ForestGreen}{(+8.0)} & 61.6 \textcolor{red}{(-4.5)} & 50.5 \textcolor{ForestGreen}{(+13.2)} & 33.9 \textcolor{ForestGreen}{(+23.3)} \\
    \midrule
    \rowcolor[rgb]{ .886,  .937,  .855} \textbf{DQRoute} & \checkmark     & \textbf{51.7 \textcolor{ForestGreen}{(+10.3)}} & {61.9 \textcolor{red}{(-4.2)}} & {52.7 \textcolor{ForestGreen}{(+15.4)}} & \textbf{38.6 \textcolor{ForestGreen}{(+28.0)}} \\
    \bottomrule
    \end{tabular}%
  \label{tab:cifar_100_lt}%
\end{table}%

\noindent \textbf{Performance on tail classes.} 
DQRoute demonstrates significant improvements on few-shot classes. On CIFAR-100-LT-IR100 (Table \ref{tab:cifar_100_lt}), DQRoute achieves \textbf{38.6\%} accuracy on the few-shot classes, outperforming prior works like SADE~\cite{zhang2022self} (33.9\%) and RIDE~\cite{wang2020long} (23.7\%). On ImageNet-LT, DQRoute obtains 43.3\% accuracy on the few-shot classes, which is competitive with SADE's best result (43.5\%), and significantly better than other methods like RIDE~\cite{wang2020long} (35.1\%) and PaCo~\cite{cui2021parametric} (39.2\%). These results clearly show that DQRoute effectively enhances recognition for underrepresented categories.

\noindent \textbf{Performance across difficulty-based methods.} 
Following previous work~\cite{son2025difficulty}, we further compare DQRoute with recent difficulty-based approaches in Table \ref{tab:difficulty}. On CIFAR-10-LT, DQRoute achieves the best accuracy across most imbalance ratios (IR), including \textbf{75.3\%} at IR 100 and \textbf{80.5\%} at IR 50. Although slightly below IDRC~\cite{yu2022re} and DBM-CE~\cite{son2025difficulty} at IR 20, DQRoute still remains highly competitive. On CIFAR-100-LT, DQRoute matches or surpasses all listed methods, achieving the best performance at IR 100 (\textbf{42.5\%}) and IR 50 (\textbf{48.3\%}), and matching the best performance (\textbf{54.7\%}) at IR 20. This confirms the robustness of DQRoute under various levels of data imbalance and difficulty.

\noindent \textbf{Overall performance across all classes.} 
In terms of overall classification accuracy, DQRoute consistently outperforms or matches the best existing methods. On CIFAR-100-LT (Table \ref{tab:cifar-100}), DQRoute achieves \textbf{51.7\%} overall accuracy at IR 100, the highest among all methods, including SADE~\cite{zhang2022self} (49.4\%) and RIDE~\cite{wang2020long} (48.0\%). DQRoute achieves 64.1\% at IR 10 and 54.9\% at IR 50, which is slightly below DBM-BSL~\cite{son2025difficulty} (65.2\%) but higher than all other baselines. On ImageNet-LT (Table \ref{tab:imagenet_lt}), DQRoute achieves an overall accuracy of 58.2\%, which is only slightly below SADE~\cite{zhang2022self} (58.8\%), and better than methods like ACE~\cite{cai2021ace} (56.6\%) and BCL~\cite{zhu2022balanced} (57.1\%). These results demonstrate DQRoute's strong generalization ability across all classes.


\noindent \textbf{Effectiveness across datasets.} 
DQRoute shows consistent improvements not only on CIFAR-100-LT, but also on ImageNet-LT and Places-LT, as shown in Table  \ref{tab:imagenet_lt} and \ref{tab:places_lt}. Its ability to generalize across diverse benchmark datasets and imbalance ratios highlights its strong practical applicability.

\begingroup
\setlength{\tabcolsep}{2pt}
\begin{table}[t]
  \centering
  \begin{minipage}{0.45\linewidth}
    \centering
     \caption{Quantitative comparison on CIFAR-100-LT. Best results are shown in \textbf{bold}, and the second-best results are shown in \underline{underline}.}
    \label{tab:cifar-100}
    \begin{tabular}{c|c|c|c}
    \toprule
    Methods & \multicolumn{3}{c}{Accuracy} \\
    \midrule
    IR    & 10    & 50    & 100 \\
    \midrule
    CE    & 59.1  & 45.6  & 41.4 \\
    BBN~\cite{zhou2020bbn}   & 59.8  & 49.3  & 44.7 \\
    BS~\cite{ren2020balanced}     & 61    & 50.9  & 46.1 \\
    RIDE~\cite{wang2020long}  & 61.8  & 51.7  & 48 \\
    LADE~\cite{hong2021disentangling}  & 61.6  & 50.1  & 45.6 \\
    ACE~\cite{cai2021ace}    & -     & 51.9  & 49.6 \\
    SADE~\cite{zhang2022self}  & 63.6  & 53.9  & 49.4 \\
    CDB-CE~\cite{sinha2020class} & 61.5  & 45.3  & 45.3 \\
    DBM-CE~\cite{son2025difficulty} & 63.1  & 51.1  & 46.5 \\
    DBM-BS~\cite{son2025difficulty} & \textbf{65.2} & \textbf{55.8} & 51.3 \\
    \midrule
    DQRoute  & \underline{64.1}  & \underline{54.9}  & \textbf{51.7} \\
    \bottomrule
    \end{tabular}%
  \end{minipage}
\hspace{1pt}
  \begin{minipage}{0.45\linewidth}
    \centering
    \caption{Quantitative comparison on ImageNet-LT. Best results are shown in \textbf{bold}, and the second-best results are shown in \underline{underline}.}
    \label{tab:imagenet_lt}
    
    \begin{tabular}{c|c|c|c|c}
    
    \toprule
    Methods & \multicolumn{4}{c}{Accuracy} \\
    \midrule
    IR    & Many  & Mid.  & Few   & All \\
    \midrule
    CE    & 68.1  & 41.5  & 14.0  & 48.0 \\
    MiSLAS~\cite{zhong2021improving} & 62.0  & 49.1  & 32.8  & 51.4 \\
    BS~\cite{ren2020balanced}    & 64.1  & 48.2  & 33.4  & 52.3 \\
    RIDE~\cite{wang2020long}  & 68.0  & 52.9  & 35.1  & 56.3 \\
    LADE~\cite{hong2021disentangling}  & 64.4  & 47.7  & 34.3  & 52.3 \\
    PaCo~\cite{cui2021parametric}  & 63.2  & 51.6  & 39.2  & 54.4 \\
    BCL~\cite{zhu2022balanced}   & 67.9  & 54.2  & 36.6  & 57.1 \\
    ACE~\cite{cai2021ace} & \textbf{71.7} & 54.6  & 23.5  & 56.6 \\
    SADE~\cite{zhang2022self}  & 66.5  & 57.0  & \textbf{43.5} & \textbf{58.8} \\
    DBM-BCL~\cite{son2025difficulty} & \underline{68.3}  & 54.3  & 38.9  & 57.6 \\
    \midrule
    DQRoute  & 64.9  & \textbf{57.1} & \underline{43.3}  & \underline{58.2} \\
    \bottomrule
    \end{tabular}%
  \end{minipage}
\end{table}
\endgroup
\vspace{-0.2cm}
\begin{table}[t]
\centering
    \caption{Quantitative comparison on Places-LT. Best results are shown in \textbf{bold}, and the second-best results are shown in \underline{underline}.}
    \label{tab:places_lt}
\begin{tabular}{c|cccccccccc}
\toprule
Methods &	CE & LFME~\cite{xiang2020learning}	& BS~\cite{ren2020balanced} &		RIDE~\cite{wang2020long} &	SADE~\cite{zhang2022self} &	PaCo~\cite{cui2021parametric} &	DQRoute \\
\midrule
Accuracy &	31.4  &	35.2 &	39.4 &	40.3 &	40.9 &	\underline{41.2} &	\textbf{41.3} \\
\bottomrule
\end{tabular}
\end{table}

\begingroup
\setlength{\tabcolsep}{9pt}
\begin{table}[t]
  \centering
  \caption{Quantitative comparison with difficulty-base methods on CIFAR-10-LT and CIFAR-100-LT. Partial results borrowed from ~\cite{son2025difficulty}. Best results are shown in \textbf{bold}, and the second-best results are shown in \underline{underline}.}
    \begin{tabular}{c|ccc|ccc}
    \toprule
    Methods & \multicolumn{3}{c|}{CIFAR-10-LT} & \multicolumn{3}{c}{CIFAR-100-LT} \\
    \midrule
    IR    & 100   & 50    & 20    & 100   & 50    & 20 \\
    \midrule
    CE    & 72.2  & 78.3  & 83.9  & 40.6  & 45.0  & 53.0 \\
    IDRC~\cite{yu2022re} & \underline{75.0}  & \underline{80.2}  & \textbf{85.5} & \underline{42.3}  & \underline{48.0}  & \underline{54.5} \\
    DBM-CE~\cite{son2025difficulty} & 74.4  & 80.0  & \textbf{85.5} & 41.1  & 46.5  & \textbf{54.7} \\
    \midrule
    DQRoute  & \textbf{75.3} & \textbf{80.5} & \underline{85.3}  & \textbf{42.5} & \textbf{48.3} & \textbf{54.7} \\
    \bottomrule
    \end{tabular}%
  \label{tab:difficulty}%
\end{table}%
\endgroup

\subsection{Ablation Studies}
\label{subsec:ablation}
\begin{figure}[htbp]
    \centering
    \begin{minipage}[t]{0.48\linewidth}
    \vspace{0.3cm}
        \centering
        \includegraphics[width=\linewidth]{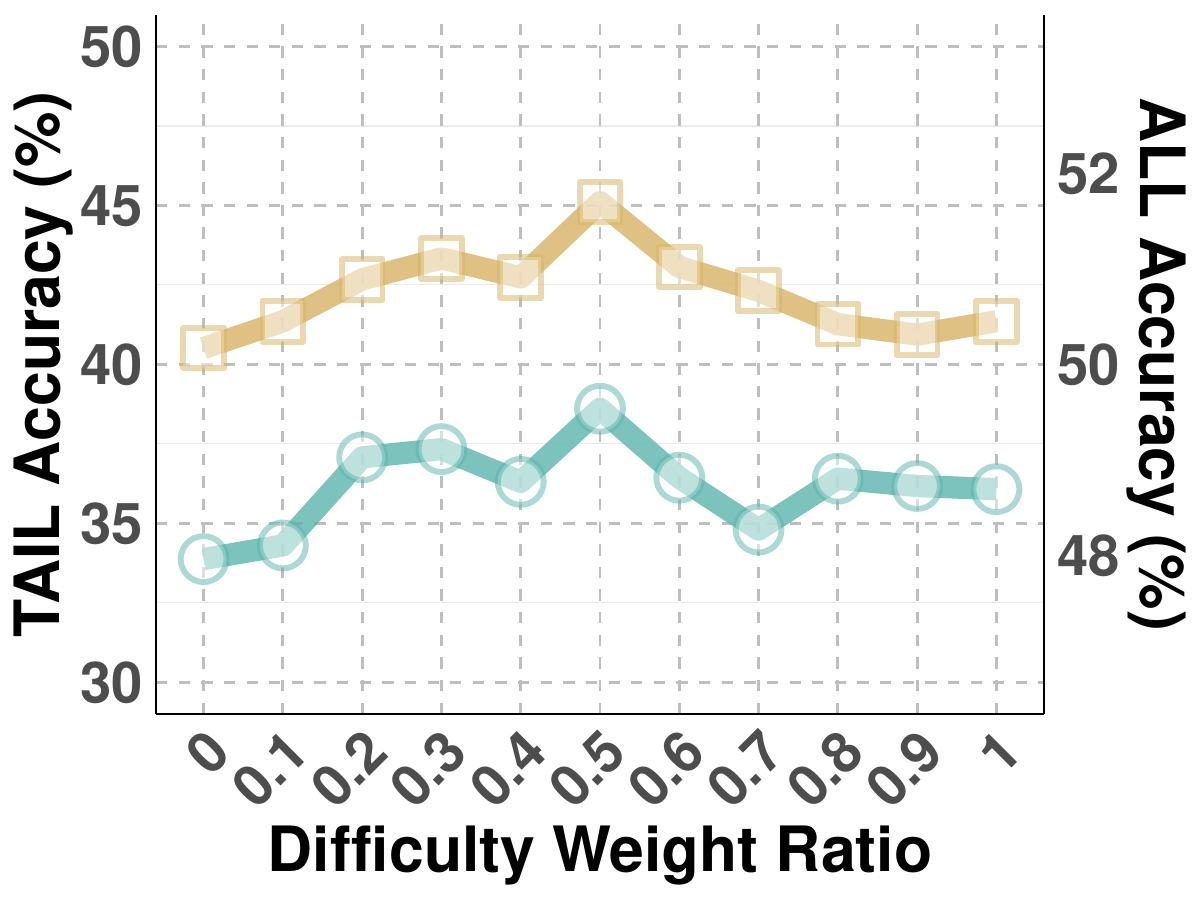}
        \caption{Ablations of different weights for difficulty.}
        \label{fig:weight}
    \end{minipage}
    \hfill
    \begin{minipage}[t]{0.48\linewidth}
    \vspace{-0.3cm}
        \centering
        \includegraphics[width=\linewidth]{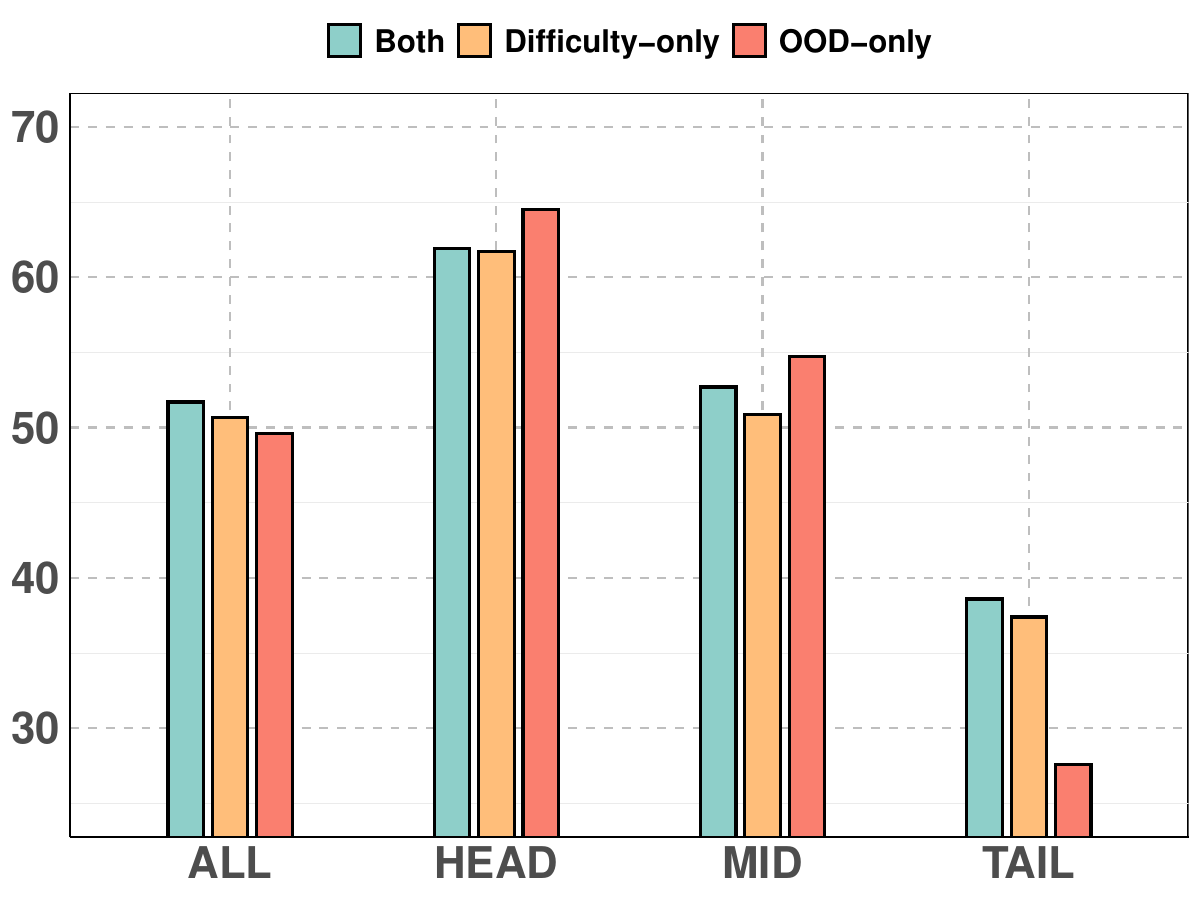}
        \caption{Ablations of different modules.}
        \label{fig:module}
    \end{minipage}
\end{figure}

\noindent \textbf{Effect of OOD Experts and Difficulty Weighting.} We conduct an ablation study to evaluate the individual and combined effects of the OOD-based expert routing and difficulty-aware reweighting components. As shown in Fig.\ref{fig:module}, using only OOD-based multiple experts improves overall performance to 49.6\%, with strong gains on head (64.5\%) and medium (54.7\%) classes but limited improvement on tail classes (27.6\%). In contrast, using only difficulty-aware weighting yields a higher overall accuracy of 50.7\%, with a substantial boost in tail performance (37.4\%), indicating its effectiveness in addressing underrepresented and ambiguous classes. Combining both components achieves the best overall accuracy of 51.7\%, along with notable improvements across all splits, especially on tail classes (38.6\%). These results demonstrate the complementary benefits of expert specialization and difficulty modeling in long-tailed recognition.

\begin{wraptable}{r}{0.5\linewidth}
\centering
\caption{Comparison of OOD loss types on CIFAR100-LT-IR100.}
\label{tab:ood_loss}
\begin{tabular}{l|cccc}
\toprule
Loss Types& All & Many & Mid. & Few \\
\midrule
BCE loss         & 49.91 & 61.69 & 52.14 & 33.57 \\
EntropyOOD       & \textbf{51.70} & 61.94 & \textbf{52.69} & \textbf{38.60} \\
Focal loss       & 50.70 & \textbf{62.74} & 50.74 & 36.70 \\
MarginOOD        & 50.45 & 61.66 & 49.57 & 38.17 \\
\bottomrule
\end{tabular}
\vspace{-1em}
\end{wraptable}

\noindent \textbf{Effect of Difficulty vs. Quantity Weighting.} We conduct an ablation study to investigate the impact of balancing difficulty and quantity in the loss reweighting strategy. As shown in Fig.\ref{fig:weight}, we vary the weighting ratio from pure quantity-based (0) to pure difficulty-based (1). The overall accuracy (ALL) improves as more emphasis is placed on difficulty, peaking at a balanced ratio of 0.5/0.5 with 51.7\%, which also yields the best tail accuracy (38.6\%). This suggests that incorporating both difficulty and quantity leads to better generalization, especially for tail classes. Overemphasis on either side results in sub-optimal performance, indicating the necessity of joint modeling.

\noindent \textbf{Effect of OOD Loss Types.} 
We compare several loss functions for training OOD experts, as shown in Table~\ref{tab:ood_loss}. Among them, EntropyOOD loss achieves the highest overall accuracy (51.70\%) and the best performance on both medium-shot classes (52.69\%) and few-shot classes (38.60\%). Focal loss performs best on many-shot classes (62.74\%), but slightly lags in few-shot classes. MarginOOD loss provides competitive results on the few-shot classes (38.17\%) but underperforms on medium classes. Overall, EntropyOOD emerges as the most balanced and effective choice across all class splits.

\section{Conclusion}
We presented \textbf{DQRoute}, a modular framework for long-tailed visual recognition. Our method combines a difficulty-aware reweighting mechanism with expert-driven dynamic routing. Specifically, we divide the label space into head, medium, and tail categories, and assign each to a specialized expert. During training, class-level difficulty is estimated based on prediction uncertainty and accuracy, and used to guide optimization. At inference time, each expert produces a prediction and a confidence score, which is used to compute routing weights for adaptive fusion. Experiments on standard long-tailed benchmarks demonstrate that DQRoute achieves consistent gains over existing methods, especially on tail categories. Future work could explore adapting the alternating-style MoE training strategy from ReconBoost~\cite{hua2024reconboost}, which effectively mitigates modality-expert competition via KL-based reconcilement regularization, to dynamically alternate training among head-, medium-, and tail-focused experts, potentially improving specialization under long-tail scenarios.

\section*{Acknowledgements}
We thank the anonymous reviewers for their helpful comments. This work is supported by the National Key R\&D
Program of China (2020AAA0107000).

\bibliographystyle{plain}
\bibliography{main}

\end{document}